# Multi-agricultural Machinery Collaborative Task Assignment Based on Improved Genetic Hybrid Optimization Algorithm

Haohao Du , Huicheng Lai, Guxue Gao, Xiaopeng Wen,

Yifei Li, Haidong Wang, Guo Zhang

[1]College of Information Science and Engineering, Xinjiang University, Urumqi 830017, China

[2]Key Laboratory of Signal Detection and Processing, Xinjiang University, Urumqi 830017, China

Correspondence: Huicheng Lai. Email: lai@xju.edu.cn

**Abstract:** Agricultural machinery collaboration technology is crucial in improving the efficiency of agricultural machinery utilisation and enhancing the benefits of large-scale production of machinery fleets. To address the challenges of delayed scheduling information, heavy reliance on manual labour, and low operational efficiency in traditional large-scale agricultural machinery operations, this study proposes a method for multi-agricultural machinery collaborative task assignment based on an improved genetic hybrid optimisation algorithm. The proposed method establishes a multi-agricultural machinery task allocation model by combining the path pre-planning of a simulated annealing algorithm and the static task allocation of a genetic algorithm. By sequentially fusing these two algorithms, their respective shortcomings can be overcome, and their advantages in global and local search can be utilised. Consequently, the search capability of the population is enhanced, leading to the discovery of more optimal solutions. Then, an adaptive crossover operator is constructed according to the task assignment model, considering the capacity, path cost, and time of agricultural machinery; two-segment coding and multi-population adaptive mutation are used to assign tasks to improve the diversity of the population and enhance the exploration ability of the population; and to improve the global optimisation ability of the hybrid algorithm, a 2-Opt local optimisation operator and an Circle modification algorithm are introduced. Finally, simulation experiments were conducted in MATLAB to evaluate the performance of the multi-agricultural machinery collaborative task assignment based on the improved genetic hybrid algorithm. The algorithm's capabilities were assessed through comparative analysis in the simulation trials. The results demonstrate that the developed hybrid algorithm can effectively reduce path costs, and the efficiency of the assignment outcomes surpasses that of the classical genetic algorithm. This approach proves particularly suitable for addressing large-scale task allocation problems.

## 1. Introduction

With the continuous advancement of science and technology, our country's mechanisation level in agriculture has significantly improved. Automation technology has become a key area in the development of modern agriculture. Agricultural machinery automatic navigation technology is crucial for implementing precision agriculture [1] and enables efficient and accurate agricultural machinery operations. Integrating farm machinery and information technology has become inevitable in developing modern agricultural machinery in China. Leveraging information technology to promote agricultural machinery development can maximise information technology's guiding role, enhance agricultural production efficiency, and hold significant importance for promoting the high-quality and efficient development of agricultural machinery in our country [2]. Smart agriculture plays a crucial role in economic development, and a range of innovations are already being applied in developed countries that enable farmers to reduce resource consumption and increase agricultural production effectively[3]. [4]Research has explored the master-slave navigation system of two agricultural machines, and experimental results demonstrate that these two agricultural machines can collaborate safely, with a 95.1% increase in efficiency compared to traditional single-machine operations. Additionally,[5]proposed a novel agricultural master-slave navigation system where the slave machine tracks the operation of the master machine, achieving preliminary collaborative navigation.

Multi-machine collaboration technology in agricultural machinery is crucial for enhancing the efficiency of machinery utilisation and improving the benefits of large-scale production of machinery fleets. Task allocation is also a significant evaluation metric for collaborative operations within agricultural machinery fleets. The challenge lies in allocating each task to the most suitable agricultural machinery unit before commencing operations, thereby maximising the overall benefits of the machinery fleet. Multi-machine collaborative task allocation has been extensively researched in logistics distribution, drones, and robotics. The integration of robotics and automation technology has significantly transformed the landscape of agriculture. The emergence of automated agriculture and the utilisation of robotics and sensors[6-9]aim to accomplish agricultural tasks more efficiently, thereby increasing production efficiency, reducing manual operations, cutting labour costs, and executing functions like field management, crop cultivation, and harvesting with greater precision. The primary goal is to address issues related to agricultural machinery's production efficiency and crop yield. Furthermore, unmanned driving technology has led to the development of high-precision autonomous driving tractors, which have applications in various agricultural activities such as ploughing and harvesting[10-11].Rational task allocation in multi-machine operations can accelerate the execution speed of fleet tasks and reduce system consumption[12], primarily aimed at minimising production costs.[13]introduced a solution utilising a dual-layer distributed Multi-Agent System (MAS) framework. They employed an iterative auction approach at the production planning layer and a distributed Hungarian method at the scheduling layer to address the multi-robot task allocation problem.[14]introduced a hybrid particle swarm optimisation and simulated annealing algorithm to find high-quality workshop scheduling solutions within reasonable computational time.[15]proposed a distributed auction algorithm based on the distance between robots and target tasks and the matching degree between robot capabilities and functions. They established a multi-robot collaborative task allocation model in a heterogeneous environment, effectively addressing the task allocation problem among heterogeneous robots. To manage the scheduling of multi-robot collaborative navigation task planning,a method based on ant colony algorithms was proposed for multi-robot collective navigation operations in agricultural fields [16]. Subsequently, a remote management platform for multi-robot coordinated navigation operations was designed and developed[17]. This platform can monitor real-time trajectories and operational information of multi-robot collaborative navigation operations and facilitate remote scheduling and management, thereby preliminary meeting the requirements of multi-robot collective navigation operations and reducing costs.[18]developed a multi-UAV task allocation algorithm based on an improved particle swarm optimisation algorithm. This algorithm builds upon the traditional particle swarm optimisation by introducing partial matching crossover and second-order swap mutation. These enhancements effectively enhance the efficiency of UAV task allocation in maritime environments and optimise navigation paths.

[19]presented a multi-UAV integrated scheduling optimisation method that divides the problem into task allocation and individual UAV scheduling stages. This approach achieves task allocation among multiple UAVs.[20]proposed a solution to the task allocation problem in multi-machine collaborative agricultural machinery operations. They devised a method based on the multi-mutated grouping genetic algorithm for assigning static tasks when similar farm machines must work together. They constructed the multi-mutated grouping genetic algorithm and designed a two-stage encoding, grouping crossover operator, and multiple mutation operators. They established a static task allocation model for multi-machine collaborative agricultural machinery operations, meeting the requirements of practical task allocation in collaborative scenarios. The research was done on multi-agricultural machinery cooperative task allocation based on an improved ant colony algorithm in agricultural field environments [21]. This was done to help manage operations involving multiple machines working together. This study laid the foundation for addressing task scheduling and management in complex agricultural operational environments that involve multi-machine collaboration.

While significant progress has been made by scholars both domestically and internationally in the field of multi-machine collaborative task allocation, there are still several limitations that need to be addressed. Firstly, a large portion of research has been applied to areas such as multi-robot systems,multi-UAVs, and multi-autonomous underwater vehicles, with only a few studies focusing on integrating agricultural machinery into the context of farming applications. Secondly, in solving task allocation problems using various algorithms, past research has mainly concentrated on optimising the algorithms' performance, such as improving efficiency and addressing issues related to local optima [22-23]. However, task allocation is a multi-constrained optimisation problem in practical agricultural field operations. Beyond minimising path costs, other factors must be considered, including supply-demand matching, agricultural machinery performance parameters, fuel consumption, and time constraints. Agricultural machinery often needs comprehensive information-gathering and scientific decision-making methods for field collaborative management. In agricultural machinery operations, there is usually an asymmetry in supply and demand information, and management authorities frequently need more scientifically rational scheduling and management plans, along with efficient and sensible task allocation strategies. Addressing these issues is essential in achieving successful multi-machine collaborative operations in agricultural fields.

Therefore, it is necessary to study the agricultural machinery static task of agricultural machinery. The innovations and contributions of this research are as follows: (1) This paper proposes a path pre-planning model based on a simulated annealing algorithm and a static task assignment model based on a genetic algorithm. (2) The tournament selection method can avoid the problem of early convergence to the local optimal solution, keep the population's diversity, and increase the algorithm's global search ability. (3) The adaptive crossover operator and adaptive mutation operator are designed. The two-segment coding method is used to optimise the objective function. A multi-crossover operator and multi-population mutation method are constructed to improve population diversity and enhance population exploration's ability to achieve multi-machine cooperative static task allocation. (4) To optimise the path further, a 2-opt local optimisation mutation operator and an improved circle algorithm are used to optimise the path better.

The rest of this paper is as follows: the second section introduces the cooperative operation model, the third section presents the task assignment model, the fourth section designs the algorithm, the fifth section provides the simulation results, and the sixth section discusses the conclusion.

## 2. Problem formulation

Task allocation refers to the process of assigning feasible tasks to suitable performers. It involves determining the nature, requirements, and priorities of jobs and giving them to appropriate individuals or resources to ensure practical task completion, maximizing the benefits obtained when all tasks are accomplished. Multi-machine collaborative static task allocation in agricultural machinery refers to establishing a mapping relationship between agricultural machinery units and multiple sections of cultivated land. Subsequently, before commencing operations, the farm machinery allocates task assignments and their order to the designated machinery units to maximize their benefits. Consequently, agricultural machinery can operate systematically in cultivated fields, facilitating coordinated scheduling and managing multiple machinery units within a specific region. Using the set $a_i = \{a_1, a_2, \cdots, a_m\}$ represents $m$ operational agricultural machinery units. Using the set $T_j = \{T_1, T_2, \cdots, T_n\}$ means $n$ tasks. The parameters of the *i-th* agricultural machinery are defined as follows $a_i = \{v_{ai}, d_i, w_i, v_i, t_{ti}, c_{vi}, c_{wi}\}, (i = 1, 2, \cdots, m)$. Table 1 lists the main parameters adopted in this study.

Table 1 the main parameters used in this study

| Symbols | Description |
|---|---|
| $v_{ai}$ | The average speed (km/h) of operation for the *i-th* agricultural machinery. |
| $d_i$ | The working width (m) of the *i-th* agricultural machinery during operation. |
| $w_i$ | The average operational capacity of the *i-th* agricultural machinery in terms of area covered (m²/h). |
| $v_i$ | The average speed (km/h) at which the *i-th* agricultural machinery travels in a non-operational state. |
| $t_{ti}$ | The average time (h) taken for the *i-th* agricultural machinery to perform a turnaround during operation. |
| $c_{vi}$ | The average fuel consumption (L/h) of the *i-th* agricultural machinery while traveling in a non-operational state per unit of time. |
| $c_{wi}$ | The average fuel consumption (L/h) of the *i-th* agricultural machinery per unit of time during operation. |
| $d_{Tj}$ | The width of the vertical operation path for task $T_j$. |
| $I_{Tj}$ | The length of the parallel operation path for task $T_j$. |
| $S_j$ | The operational area of task $T_j$. |

For analysis, a section of farmland was selected in the farm testing field. This area was divided into 15 idealized plots based on the requirements. Each parcel of farmland was considered an obstacle, and free traversal was restricted due to the distribution of the farmland. The *Ovitalmap* was utilized to establish the farmland model, and the model was visualized. In the model, white indicates roads, and blue represents boundaries, as depicted in Fig 1.

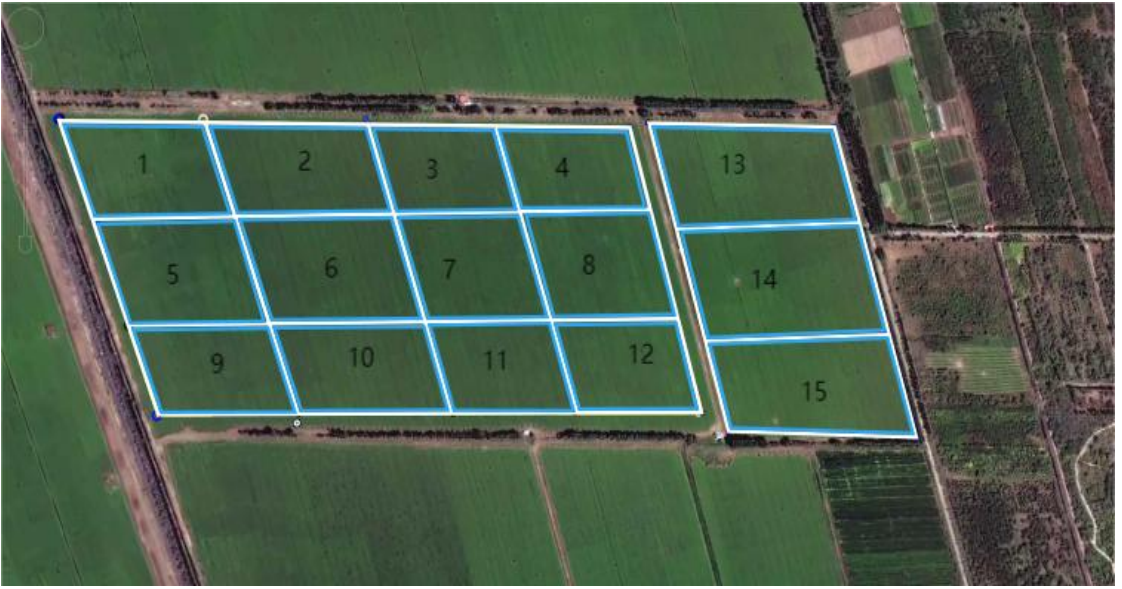

Fig.1 Model of experimental farmland

**2.1.The primary hypothesis of multi-machine cooperative operation**

Due to the complexity of collaborative operations and to simplify the problem and facilitate computation, the following basic assumptions are made based on standard operational scenarios: ① The performance parameters of agricultural machinery and the positional parameters of farmland are known. Each plot of land is not connected to any other plot.② The number of tasks must be greater than the number of farm machines.③ Each task plot is assigned to only one agricultural machinery unit, with the route planning for each task plot known.④ Agricultural machinery departs from the garage, completes all assigned tasks, and returns to the garage.⑤ The farmland is obstacle-free. The agricultural machinery is in an ideal state.

**2.2.Multi-machine cooperative operation model**

(1) The cost of completing a task for an agricultural machinery unit is calculated as the distance the machinery travels from the garage to the designated task location and back to the garage. Each agricultural machinery unit's overall operational time and fuel consumption are computed. In practical applications, agricultural machinery typically departs from the garage to sequentially visit task areas, each being called only once. The machinery returns to the garage after completing the final task. The cost associated with the path of each machinery unit is as follows:

$$s_i = D^i_{s^i_0, T^i_1} + \sum_{k=1}^{T^i-1} D(T^i_k, T^i_{k+1}) + D^i_{s^i_0, T^i_{n^i}} \tag{1}$$

Formula 1.Where $s_i$ represents the path cost of agricultural machinery $i$.Where $D^i_{s^i_0, T^i_1}$ represents the distance from the starting point $s_i$ to the first task point $T^i_1$ for agricultural machinery $i$.Where $D^i_{T^i_{n^i}, s^i_0}$ represents the distance from the last task point $T^i_{n^i}$ to the return point $s^i_0$ for agricultural machinery $i$ .Where $\sum_{k=1}^{T^i-1} D(T^i_k, T^i_{k+1})$ represents the total distance traveled by agricultural machinery $T^i_1$ from task e to $T^i_k$.The distance is calculated as shown in Formula (2).

$$D(T_k, T_{k+1}) = \sqrt{(x_{k+1} - x_k)^2 + (y_{k+1} - y_k)^2} \tag{2}$$

The objective function is established to minimize the total path cost for all agricultural machinery units,and it is defined as follows:

$$\min f_{总} = \min\left(\sum_{i=1}^{M}\left(D^i_{s^i_0,T^i_1} + \sum_{k=1}^{T^i-1} D(T^i_k, T^i_{k+1}) + D^i_{T^i_{n^i},s^i_0}\right)\right) \quad (3)$$

$$\sum T^i = N, \quad i \in M \quad (4)$$

Formula (3) represents the minimization of the total path cost between task areas,while Formula (4) represents that all agricultural machinery units need to complete tasks in all task areas.

The fuel consumption $c_i$ comprises three parts: fuel consumption during travel,fuel consumption for turning during operation,and fuel consumption during operation.This is illustrated in Formula (5).

$$c_i = \frac{s_i}{v_i} * c_{vi} + \sum_{j=1}^{n} n\_turn(ij) * t\_T(i) * c_{vi} + \sum_{j=1}^{n} \frac{S_i}{w_i} * c_{wi} \quad (5)$$

In Formula (5),*n_turn(ij)* represents the number of U-turns for the *i-th* agricultural machinery at the *j-th* task location.

(3)The total time $t_i$ to complete tasks consists of three components: machinery travel time,machinery in-field turning time,and machinery operation time.This is represented by Formula (6).

$$t_i = \frac{s_i}{v_i} + \sum_{j=1}^{n} n\_turn(ij) * t\_T(i) + \sum_{j=1}^{n} \frac{S_i}{w_i} \quad (6)$$

## 3. Establishment and Process Design of Multi-Machinery Collaborative Task Allocation Model

### 3.1.Framework for Task Allocation of Multiple Agricultural Machinery

The collaboration of various agricultural machinery can be divided into two categories: static task allocation and dynamic task allocation. This article mainly focuses on the static task allocation for multiple agricultural machinery working collaboratively, as illustrated in Fig.2.In static task allocation, tasks are reasonably allocated to each agricultural machine, and an efficient and rational global resource allocation plan is generated based on the known task information.

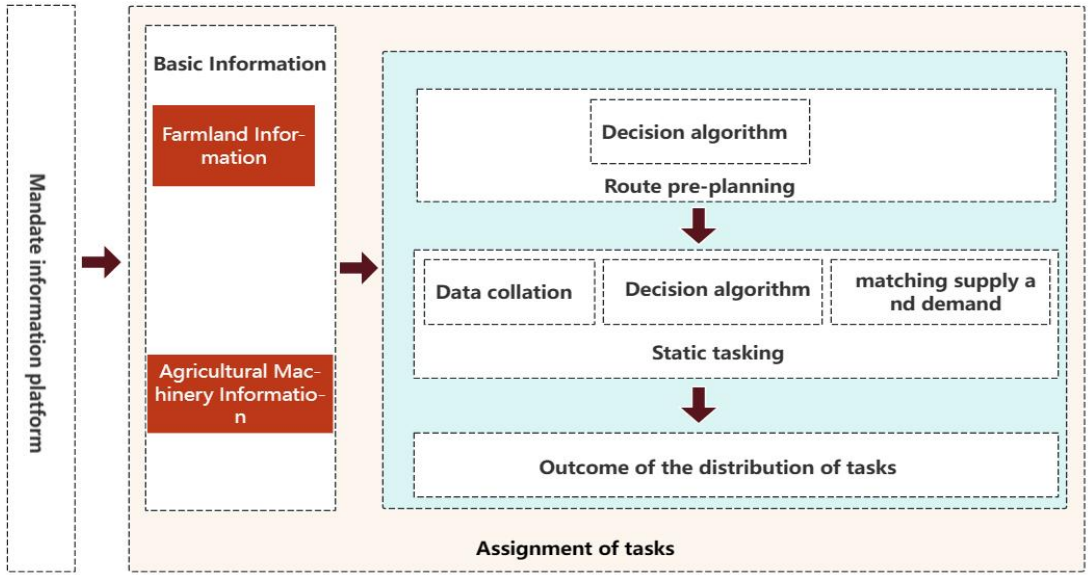

Fig.2 Overall framework of multi-computer cooperative task allocation relationship

### 3.2 Task Allocation Methods

There are $\frac{n!(n-1)!}{(m-1)!(n-m)!}$ possibilities to assign n tasks to m-machine in an orderly way.When n is

large,we can not use the exhaustive method to get the optimal task assignment.At this time,we often use the heuristic method to get a feasible solution [24-25];in this paper,a simulated annealing algorithm and a genetic algorithm are used to solve the problem.

**3.2.1 Path Preplanning Based on Simulated Annealing Algorithm**

Simulated Annealing is a global optimisation algorithm[26]used to find the optimal solution to a problem within a search space. Inspired by solid-state annealing principles, it simulates a solid object's cooling process from high to low temperature to explore the solution space. This paper employs the Simulated Annealing algorithm for path preplanning to obtain a feasible solution with the lowest cost. The specific process is outlined below:

(1) Set the initial value of the iteration count,initial temperature $t_f$ ,final temperature $t$,and the length of the Markov chain.

(2) Initialize the current offspring population as *route_new=route* and set the initial distance matrix and path.

(3) Sets the optimal path length to infinity and initializes the current and optimal paths to the initial directions.

(4) When the temperature is greater than or equal to the last temperature, perform the following steps:

① The number of iterated Markov chain lengths

② Perform path perturbation to generate a new path route.

③ Calculates the size of the new path.

④ If the length of the new way is less than the length of the current path, the length of the current path is updated to the length of the new path.

⑤ Otherwise, the current new path is updated with a certain probability according to Metropolis criteria. If the new path length is less than the optimal path length, the optimal path and length are updated.

(5) Update the temperature.

(6) Finally, the optimal path and length are returned as the algorithm's output.

**3.2.2.Task Allocation Based on Genetic Algorithm**

Genetic Algorithm (GA) is a heuristic optimisation algorithm [27] inspired by the principles of biological evolution and genetics in nature. It simulates the process of natural growth and has strong global search capabilities, making it effective in finding better solutions in large search spaces. Genetic algorithms typically involve three basic steps: selection, crossover, and mutation. An adaptive crossover operator and a genetic algorithm with multiple adaptive modifications are proposed. This algorithm employs two-stage encoding and uses tournament selection of chromosomes, effectively addressing the problem of multi-machine collaborative task allocation.

**4. Algorithm Design**

**4.1.Encoding**

According to reference[28], a two-stage encoding method was employed, as illustrated in Fig.3.The first part represents the task order, with *n* tasks represented by numbers from 1 to *n*, utilising *n* digits. The second part denotes the number of tasks assigned to each agricultural machine (*AM*) and consists of *m* digits. For instance, the encoding scheme is applied when *n* = 8 and *m* = 3.

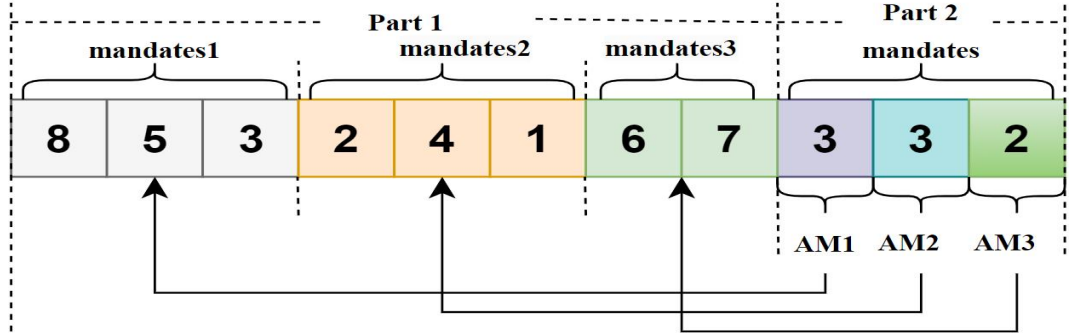

Fig.3 Two-stage coding

**4.2. Selection**

Tournament selection is simple, effective, and easy to implement. It involves selecting a small group of individuals to compete and selecting the fittest as the winner. This selection method has a low computational cost and is suitable for large-scale problems and complex optimisation tasks. Additionally, it aids in maintaining diversity; tournament selection allows weaker individuals to participate in the competition, increasing the diversity of the selection process. Even if an individual has relatively low fitness, he still has the chance to be selected, thus obtaining the opportunity for survival and reproduction. This helps avoid the problem of early convergence to local optima, maintains population diversity, and enhances the algorithm's global search capability.

Define the problem's fitness function, where higher-fitness chromosomes represent better task allocation outcomes. The fitness function is defined as shown in Formula (7):

$$F_{fit} = \frac{1}{f} \tag{7}$$

**4.3 Adaptive Crossover Operator**

(1) The sigmoid function is a commonly used logistic function, also called the logistic function. It maps real numbers to the interval (0,1) and is frequently utilized to convert continuous inputs into probability values. Its formula is given by Formula (8):

$$F_{fit} = \frac{1}{1+e^{-x}} \tag{8}$$

Where x is the input value, e is the base of the natural logarithm (approximately equal to 2.71828). The output of the function is a value between 0 and 1.

(2) In Formula (10). Using the method of adaptive step size, we calculate a relative evolutionary algebraic value according to the number of population (*N*) so that its range *r*, Formula (9) is shown. The step size of *R* will increase with algebra, and the value will rise first and then decrease. This scaling effect can be used to control the genetic algorithm's early exploring ability and later balancing power; this makes evolution more exploratory in the early stages and more balanced in the later stages so that fitness changes over a smaller range. The R trend is shown in Fig.4.

$$r = 2 * log(N) * \frac{gen}{MAXGEN} - log(N) \tag{9}$$

$$R = \frac{1}{1+exp(-r)} * \frac{(MAXGEN-gen+1)}{MAXGEN} \tag{10}$$

Where *n* is the number of population, *Gen* is the number of current iterations, and *MAXGEN* is the total number of iterations.

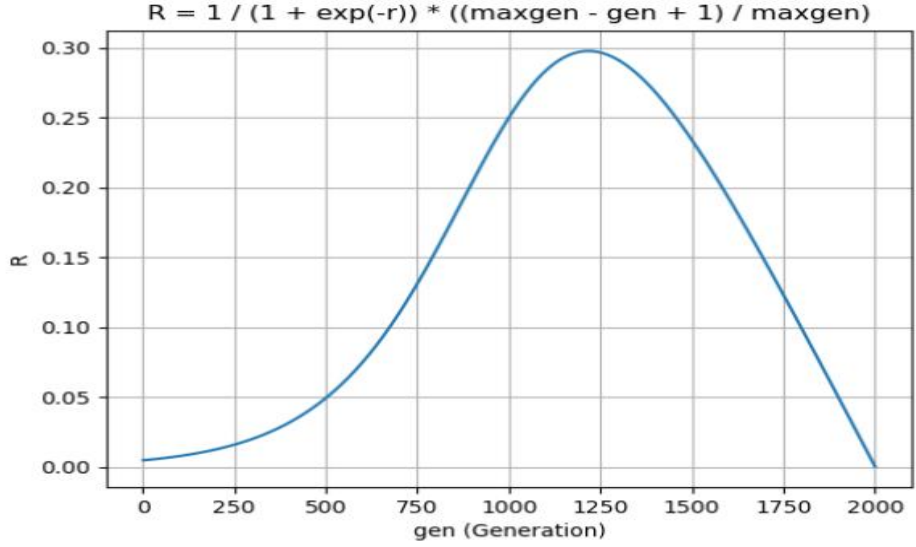

Fig.4 Trends in R

(3) Overshoot

$$overshoot = \frac{(Max(D_k) - Mean(D_k))}{Mean(D_k)} \quad (11)$$

In Formula (11), $max(D_k)$ represents the maximum value of the path cost for agricultural machine $m(i)$, and $Mean(D_k)$ represents the average path cost for all agricultural machines $m$.

$$to1 = \frac{(Max(D_{k1}) - Mean(D_{k1}))}{Mean(D_{k1})} \quad (12)$$

$$to2 = \frac{(Max(D_{k2}) - Mean(D_{k2}))}{Mean(D_{k2})} \quad (13)$$

In Formula (12)-(13), *to1* and *to2* represent the two individuals' overshoot.

(4) Definition of adaptive tolerance as shown in Formula (14): the crossover operator can be dynamically adjusted, and in the global search phase, the population diversity can be increased by increasing the crossover probability to explore the solution space better. In the local search phase, the crossover probability can be reduced gradually, which makes the algorithm search the local area near the excellent solution more centralized.

$$sp = tolmin + R * \frac{gen*(tol_{max} - tol_{min})}{Maxgen} \quad (14)$$

*Tolmin* and *Tolmax* are two given thresholds representing the parameter's minimum and maximum values, respectively. *R* is an adaptive step size to achieve dynamic adjustment and optimization of parameters.

(5) Based on the new population resulting from the selection of the bidding tournament, two chromosomes are randomly selected in the new population, and the crossover probability is calculated separately for each chromosome. As shown in Formula (15) - (16).

$$P_{c1} = P_{cmax} * (y1\_mean - y1\_max) + P_{cmin} \frac{y1\_mean - y1\_min}{y1\_mean - y1\_min} \quad (15)$$

$$P_{c2} = P_{cmax} * (y2\_mean - y2\_max) + P_{cmin} \frac{y2\_mean - y2\_min}{y2\_mean - y2\_min} \quad (16)$$

Where *Pcmax* and *Pcmin* denote two given crossover probabilities, respectively, The average, maximum, and minimum path costs of each agricultural machine among the *m* agricultural machines in the first chromosome are represented by *y1_mean*,*y1_min* and *y1_max*.
The crossover probabilities of the two randomly selected chromosomes are normalised to obtain the crossover probabilities. To ensure that each possibility is correctly considered and applied, the problems associated with inconsistent probability distributions are avoided, thus better achieving the desired effect of the optimisation algorithm. As shown in Formula 17-18.

$$P_{c2} = P_{c1} * w_1 + P_{c2} * w_2 \tag{17}$$
$$w_1 + w_2 = 1 \tag{18}$$

**4.3.1 Sequential intersections (OX)**

The crossover process of the sequential crossover operator is shown in Fig.5 ① Select two individuals (*Dad* and *Mom*) randomly.② Randomly select two crossover points (*CrossPoint1* and *CrossPoint2*),such that *CrossPoint1* is smaller than *CrossPoint2*.e.g., *CrossPoint1*=2, *CrossPoint2*=4.③ The mother's chromosome is passed as it is in the gene section between *CrossPoint1* and *CrossPoint2*.④ The remaining partial gene segments in *Mom* are inserted into the remaining segments of the offspring (*offspring*) in the order in which they appear in *Dad*. A random breakpoint is selected for the second part of the chromosome, and then the two gene segments are reversed in order.

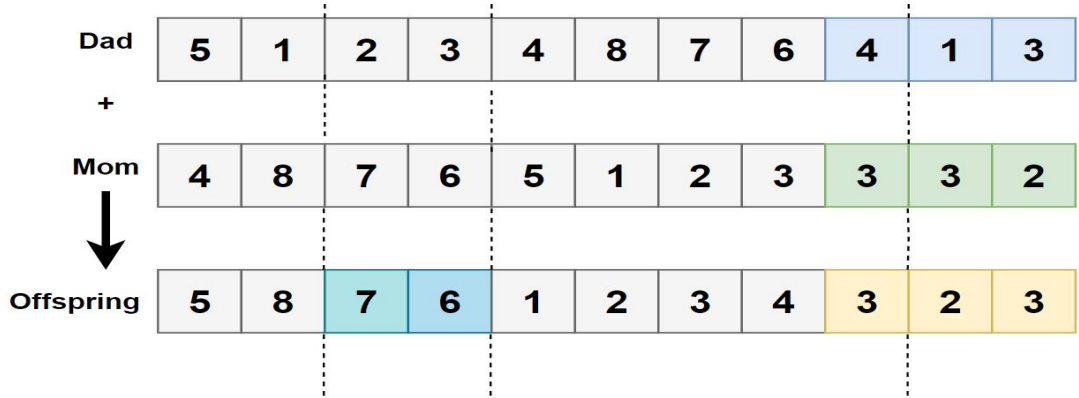

Fig.5 Sequential crossover operator

**4.3.2 Partially-Mapped Crossover (PMX)**

The crossover process of the partially mapped crossover operator is shown in Fig.6. ① Two chromosomes are randomly obtained from the selected population.② In the first part of the chromosome, two random numbers are generated to satisfy 0≤*k1*<*k2*≤ *n* the length of the chromosome, such as *k1*=2,*k2*=6, as the starting position of the intercepted chromosome segments.③ The two intercepted segments are exchanged in position, and the new offspring can be obtained.④ The final zygote chromosome can be obtained by filling it according to the mapping relationship.⑤ In the second part of the chromosome, a random crossover point is generated, and then the two gene segments are reversed.

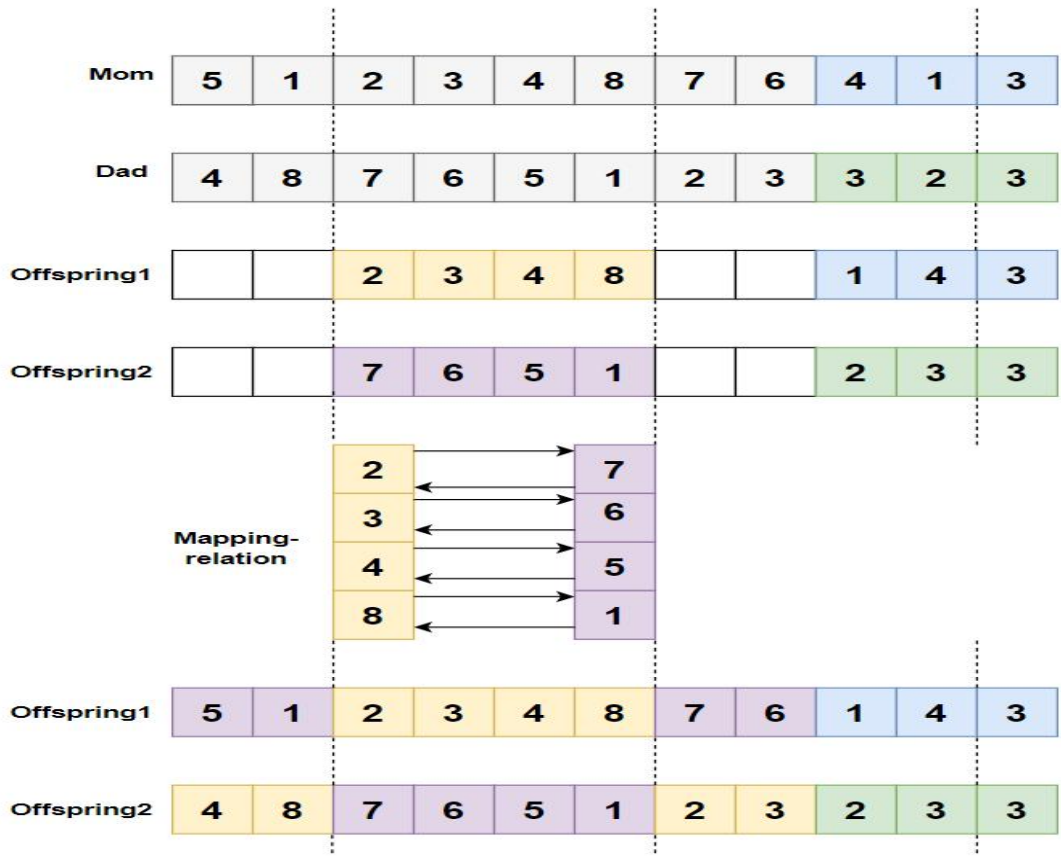

Fig.6 Partial mapping crossover operator

The design process of the adaptive crossover operator is shown in Fig.7, which is implemented as follows: ① Use the tournament selection method to get a new population to select two chromosomes randomly; one is the parent generation, the other is the mother generation, the number of newly generated individuals must be even, if it is odd will be converted to an even number of individuals in a certain way.② The crossover probabilities $P_{c1}$ and $P_{c2}$ were calculated for the paternal and maternal generations,respectively,and the total crossover probability $P_c$.③ The random number $R_r \in [0,1]$,generated，If $R_r<P_m$,perform crossover.Then,assess the magnitudes of tolerance (*sp*) and the parent's overshoot(tol).If *to1<sp*,apply ordered crossover (*OX*) operation; otherwise,use partially-mapped crossover (*PMX*) operation.Apply the same logic to the maternal parent.④ Remove the randomly selected chromosome from the population,then repeat the process of step ③ With the remaining individuals by randomly selecting two individuals again.⑤ Generate a new population.

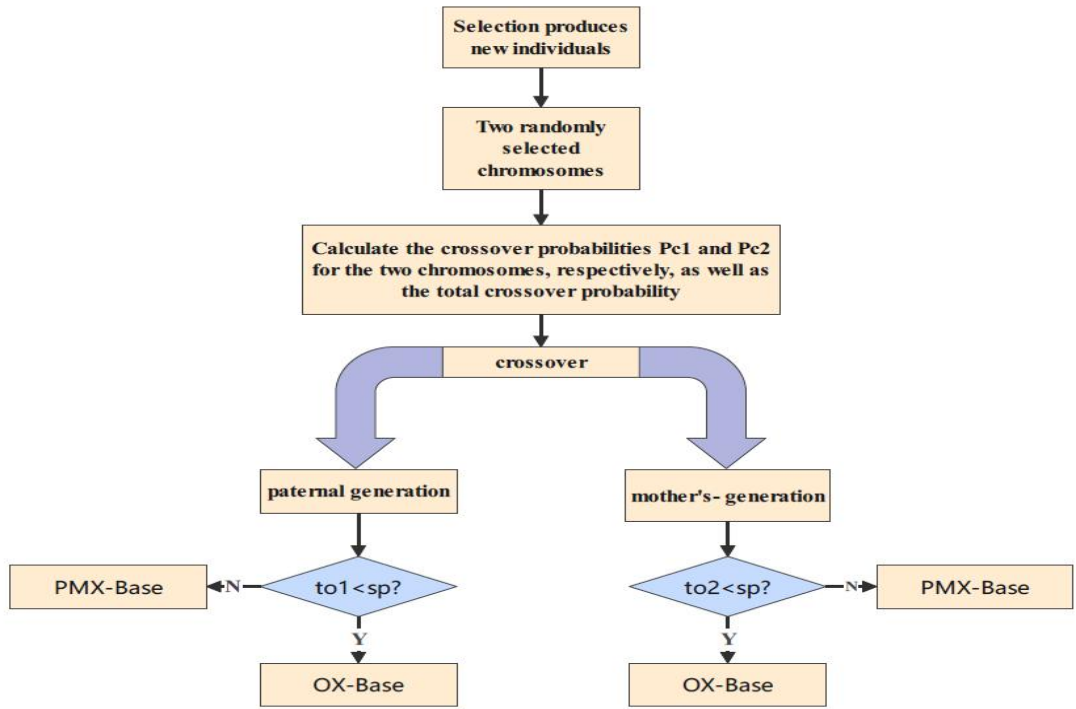

Fig.7 Adaptive intersection operator

**4.4.Mutation operator**

**4.4.1.Exchange mutation operator**

The process of exchanging variation operators is shown in Fig.8,① Select an individual (*Dad*) randomly.② Two gene positions (Position 1 and Position 2) are randomly selected in the first part of the chromosome. Their values are exchanged.③ A new individual (*the offspring*) is produced. The exchanged genes are placed in the offspring.④ Generate two random crossover points in the second part of the chromosome and exchange the gene fragments.

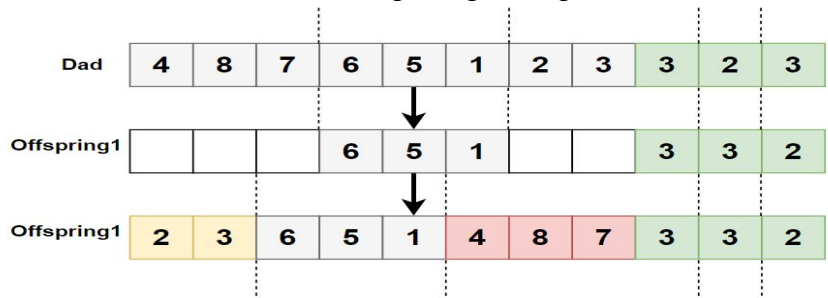

Fig.8 Exchange of variational operators

**4.4.2 Insert Mutation Operator**

Inserting the mutation operator is shown in Fig.9,① Randomly select an individual (*Dad*).② Two gene positions (*Position 1* and *Position 2*) are randomly selected in the first part of the chromosome.The gene from Position 2 is inserted into the position before Position 1.③ A new individual (*offspring*) is generated,and the inserted gene is placed in the offspring.④ Generate two random crossover points in the second part of the chromosome for gene fragment insertion.

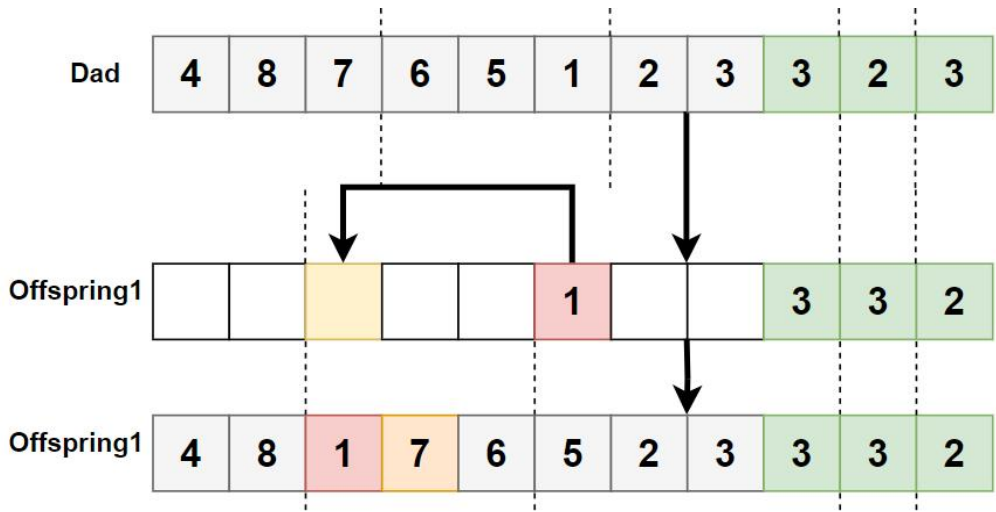
Fig.9 Insertion of mutation operator

4.4.3 The process of designing the adaptive mutation operator is illustrated in Fig.10, and its implementation is as follows:① Calculate the similarity of each individual.② Calculate the diversity of each individual.

$$div = sum(sim, 2)./(pop\_size - 1) \quad (19)$$

In Formula 19, *sim* denotes the value of similarity, *pop_size* denotes the size of the population, and *div* characterises the diversity of each individual.③ Adjust the mutation rate based on the fitness value,as shown in Formula (20).

$$P_m = (1 - fit) * p_m + div * fit \quad (20)$$

Where $p_m$ represents the initial mutation probability, $P_m$ indicates the adaptive mutation probability, and fit represents the fitness value of each individual in the population.④ Perform mutation based on the mutation rate. If $R_r<P_m$ is satisfied, perform crossover. Randomly select a mutation method, apply it to the *i-th* individual, and generate a new population.

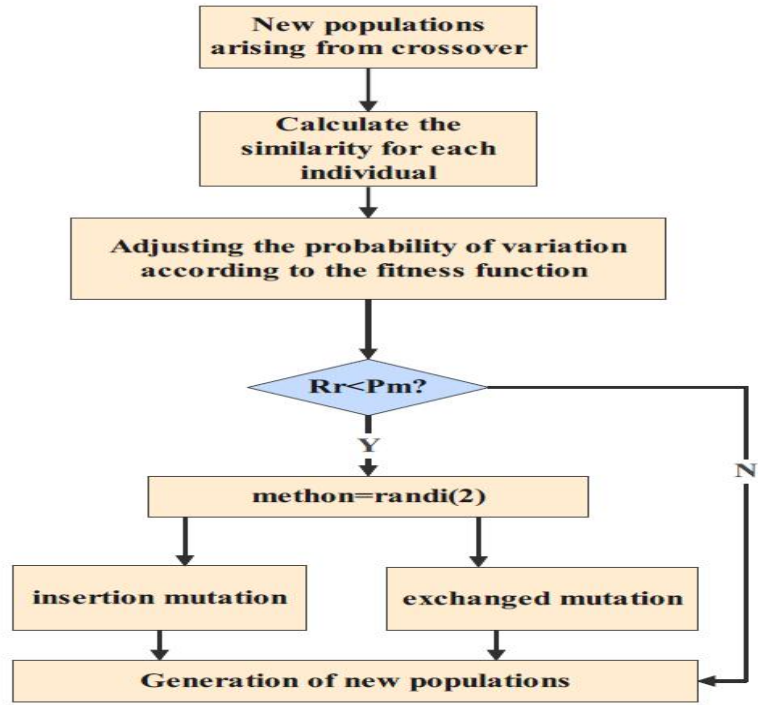
Fig.10 Adaptive mutation operator

**4.5.2-opt Optimization of local variation optimizer and Circle modification algorithm**

(1) 2-opt belongs to the local search algorithm, which is an effective tool for solving combinatorial optimisation problems. The algorithm is implemented by ① Randomly selecting 2 points, *i* and *j*, in the chromosome produced after mutation.② The path before *i* is unchanged and added to the new way; the path between *i* and *j* is flipped in its encoding and added to the new approach; and the path after *j* is unchanged and added to the new way.③ The second part is the same.

The schematic diagram of the 2-opt algorithm is shown in Fig.11, which shows that the gene grouping breakpoints in the first part are $i = 2, j = 6$, and the gene grouping breakpoints in the second part are $i = 1, j = 2$.

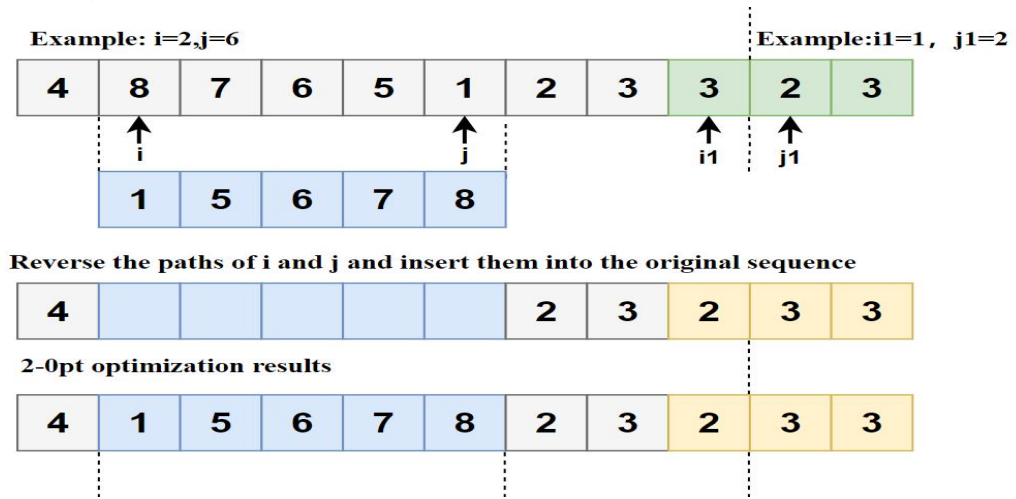

Fig.11 2-opt locally optimized variational operator

(2) Modified circle algorithm is an optimization algorithm, thus making it better in every aspect and can provide better performance, accuracy and adaptability. First, a Hamilton circle is obtained, then the process is modified to get another Hamilton circle with lesser weights until it cannot be improved, and then it stops. For $1 \leq i < i+1 < j \leq n$,(2) Construct a new Hamilton circle, described as shown in Fig.12.

| Algorithm | Circle Modification Algorithm |
|---|---|
| Input: | Initialise the input data and variables and calculate the distance matrix between individual farm fields D. |
| Output: | Output the optimal solution sequence. |
| Step 1 | Create an initial circle *c*, representing the initial farm path. *c* is a vector containing the farm indexes. |
| Step 2 | Initialise the exit flag flag to 0. |
| Step 3 | In a double loop, iterates over each pair of neighbouring cities (*i*,*j*) of the city path, checking if exchanging these two paths reduces the total path length. |
| Step 4 | Determine whether a relatively optimal solution has been found based on the value of the exit flag. If the flag is still 0, it means that there are no swaps that can be optimised, and a relatively optimal solution has been found. Calculate the path length by accumulating the distance from the starting point through all intermediate farms and back to the starting point. |
| Step 5 | Store the current path c as the final solution circle. |

Fig.12 Circle Modification Algorithm

### 4.6 Algorithmic flow

The flow of the designed algorithm in this paper is shown in Fig.13.

(1)Initialize the population according to the problem and use the SA algorithm for path preplanning.

(2) Based on the new population generated by the SA algorithm, if a new population is generated, then continue to use the genetic algorithm to randomize the task assignment for each individual; otherwise, repeat step (2) until the last individual is executed.

(3) Select the better-adapted individual based on the tournament approach.

(4) According to the coding method, randomly select two individuals in the chosen population, delete the randomly selected individuals, perform crossover based on the crossover method, calculate the adaptation value of each individual to find the best individual, and then complete the mutation operation on the new population based on the process of adaptive mutation until the new population is generated, calculate the adaptation value of each individual, and compare the optimal produced by crossover and mutation fitness values to find the optimal solution.

(5) Optimize the optimal solution using 2-opt and modified complete circle algorithms to obtain the new optimal individual. If the current number of iterations (*gen*) is less than the maximum number of iterations (*MAXGEN*), repeat steps (3)-(5).

(6) If the number of iterations is reached, the individual with the highest fitness of the current population is selected as the solution for this task assignment.

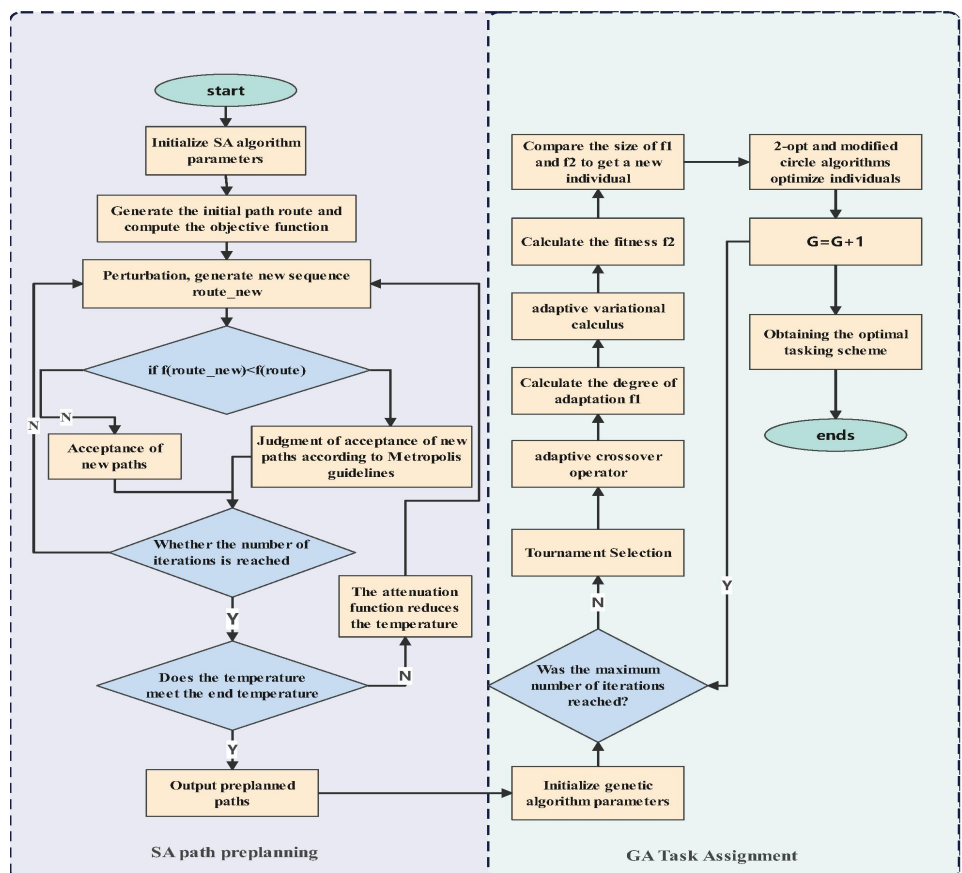

Fig.13 Flow of the algorithm

## 5 Algorithm Simulation and Data Analysis

### 5.1 Optimization Performance Testing of Hybrid Algorithms

In the simulation experiments based on hybrid algorithms, to validate the effectiveness of the proposed approach, several hundred simulation experiments were conducted using MATLAB R2020b on a computer with an Intel(R) Core(TM) i7-8750H CPU running at 2.20 GHz and 8GB RAM.

Detail settings: Initial temperature $t$=120, final temperature $t_f$=1, the number of populations $N$=200, *Markov* chain length is set to 100; the maximum number of iterations *MAXGEN* is set to 2000, the initial crossover probabilities of $P_{cmin}, P_{cm}, P_{cmax}$ are 0.4, 0.55, 0.85, respectively, the $P_m$ initial variance probability is 0.1. The *tolmax,tolmin* are set to 0.8, 0.05.

**5.2 MTSP experimental comparison**

In this experiment, we conducted a series of trials on multitasking task allocation, covering algorithms such as IPGA and GA2PC [29], GA, GAG, HCGA and SAGA[30-31], GA-PMX, GA-OX and GA-CX[32], as well as the improved genetic hybrid optimization algorithm that we proposed. To evaluate the performance of the hybrid algorithms and our approach, we use multiple datasets publicly available in TSPLIB, including eil51, eil76, eil101, kroA100, kroA150, kroA200, att48, lin318, rat575, pr1002, U1432, vm1748, pcb3038, etc. On different datasets, we conducted several experiments and calculated the average values to derive the average distance cost, and the relevant data are shown in Tables 2,3, 4 and 5. Compared with other methods, our improved genetic hybrid optimization algorithm performs well. It is the best performer regarding the total distance metric, which can effectively reduce the path cost. These results further demonstrate the superior performance of our improved genetic hybrid optimization algorithm in the multitasking task allocation problem.

Table 2 Comparison of distance cost between IPGA,GA2PC and improved genetic hybrid optimization algorithms

| No. | case | m | IPGA | GA2PC | IGHOA |
|---|---|---|---|---|---|
| 1 | Eil51 | 3 | 478 | 543 | **471.77** |
| 2 | | 5 | 549 | 586 | **506.13** |
| 3 | KroA100 | 3 | 24288 | 26653 | **25041.00** |
| 4 | | 5 | 30710 | 30408 | **31147.00** |
| 5 | KroA150 | 3 | 35428 | 47418 | **33606.00** |
| 6 | | 5 | 43725 | 49947 | **39063.00** |

Table 3 Comparison of distance cost of GA,GAG,HCGA,SAGA and improved genetic hybrid optimization algorithm

| No. | case | m | GA | GAG | HCGA | SAGA | IGHOA |
|---|---|---|---|---|---|---|---|
| 1 | Eil51 | 3 | 657.97 | 509.99 | 476.30 | 472.56 | **471.77** |
| 2 | | 4 | 569.20 | 549.20 | 522.97 | 503.34 | **487.28** |
| 3 | | 5 | 624.18 | 608.55 | 581.89 | 567.32 | **506.13** |
| 4 | Eil76 | 3 | 980.54 | 666.80 | 661.20 | 653.96 | **610.98** |
| 5 | | 4 | 849.18 | 695.06 | 694.64 | 690.94 | **635.35** |
| 6 | | 5 | 1018.18 | 780.27 | 771.50 | 747.34 | **627.57** |
| 7 | | 6 | 862.45 | 825.35 | 804.21 | 800.84 | **642.42** |
| 8 | Eil101 | 4 | 1092.61 | 817.89 | 803.38 | 796.96 | **772.50** |
| 9 | | 5 | 1137.01 | 847.48 | 857.54 | 867.24 | **770.23** |
| 10 | | 6 | 1099.08 | 75.62 | 900.34 | 900.90 | **803.34** |
| 11 | | 7 | 1062.69 | 900.97 | 898.77 | 900.72 | **803.11** |

Table 4 Comparison of distance cost of GA,HCGA and and improved genetic hybrid optimization algorithms

| No. | case | m | GA | HCGA | IGHOA |
|---|---|---|---|---|---|
| 1 | Eil101 | 4 | 921.2 | 813.0 | **772.50** |
| 2 | Lin318 | 4 | 135188.2 | 71646.4 | **49237.60** |
| 3 | Rat575 | 4 | 33008.9 | 13057.0 | **8541.41** |
| 4 | Pr1002 | 3 | 3218723.7 | 721022.8 | **305517.00** |
| 5 | Pr1002 | 6 | 1524822.4 | 538609.8 | **317816.00** |
| 6 | U1432 | 3 | 2072059.0 | 483942.8 | **170505.00** |
| 7 | U1432 | 6 | 914170.0 | 308176.0 | **171523.00** |
| 8 | Vm1748 | 3 | 7719677.9 | 1858539.5 | **609624.00** |
| 9 | Vm1748 | 6 | 3365320.0 | 938670.90 | **697024.00** |
| 10 | Pcb3038 | 3 | 3699627.9 | 1302418.6 | **163582.00** |
| 11 | Pcb3038 | 4 | 1837962.6 | 657501.3 | **163248.00** |
| 12 | Pcb3038 | 6 | 1497143.3 | 427950.6 | **165281.20** |

Table 5 Comparison of distance cost of GA-PMX,GA-OX,GA-CX and improved genetic hybrid optimization algorithms

| No. | case | m | GA-PMX | GA-OX | GA-CX | IGHOA |
|---|---|---|---|---|---|---|
| 1 | Att48 | 2 | 52343.48 | 52023.49 | 52773.84 | **37752.60** |
| 2 |  | 4 | 64954.87 | 66324.27 | 64803.34 | **37514.20** |
| 3 |  | 6 | 75357.03 | 78344.59 | 77401.42 | **40504.60** |
| 4 |  | 8 | 93423.03 | 96930.30 | 94066.53 | **44422.40** |
| 5 | KroA100 | 2 | 32145.24 | 32872.40 | 32423.66 | **24948.00** |
| 6 |  | 4 | 46748.49 | 48820.87 | 47549.87 | **29006.00** |
| 7 |  | 6 | 57152.93 | 59349.57 | 56772.45 | **26652.00** |
| 8 |  | 8 | 58265.54 | 65924.12 | 62509.56 | **28524.70** |
| 9 | KroA150 | 2 | 43113.54 | 43381.04 | 42835.79 | **32203.00** |
| 10 |  | 4 | 59332.85 | 61113.17 | 61185.34 | **36903.00** |
| 11 |  | 6 | 71069.14 | 74566.69 | 72132.12 | **35202.00** |
| 12 |  | 8 | 81819.18 | 86174.12 | 84449.72 | **35568.70** |
| 13 | KroA200 | 2 | 48368.55 | 46622.32 | 48537.27 | **36763.30** |
| 14 |  | 4 | 66758.95 | 67152.20 | 69299.25 | **36669.10** |
| 15 |  | 6 | 82791.72 | 82620.29 | 84913.85 | **38508.00** |
| 16 |  | 8 | 95258.01 | 97601.72 | 99075.72 | **39036.50** |

**5.3 Simulation and Analysis**

To verify the effectiveness of the hybrid algorithm in task allocation, we designed a set of simulation experiments involving task-area allocation. In the experiments, we used different numbers of farm machines (3, 5 and 6) for task allocation. We used the performance parameters of each of the three farm machines according to the data provided in the reference[20]. The performance parameters of the farm machines are shown in Table 6, and the total number of tasks is 16, with the main parameters of each job shown in Table 7. These task fields are relatively regular rectangular fields. We evaluate the effectiveness of the improved genetic hybrid algorithm by comparing the performance of the traditional and enhanced genetic hybrid algorithms on the static task allocation problem. Through these experiments, we aim to verify whether the improved hybrid algorithm can better solve the task allocation problem to improve the efficiency of farmland operation and maximise the overall benefits.

Table 6 Performance parameters of agricultural machinery

| Agricultural machine serial number | Operating width/m | Operating capability/($m^2$/h) | Road speed/(km/h) | Fuel consumption in operating condition/(L/h) | Non-operating fuelconsumption/(L/h) | Turnaround time/h |
|---|---|---|---|---|---|---|
| 1 | 6.9 | 7000 | 10 | 7 | 3.0 | 0.004 |
| 2 | 5.5 | 5000 | 10 | 5 | 2.5 | 0.003 |
| 3 | 3.7 | 4000 | 10 | 4 | 2.0 | 0.002 |

Table 7 Parameters of the mission field

| Task number | Lengths/m | Widths/m | Area /$m^2$ |
|---|---|---|---|
| 1 | 70 | 127 | 8890 |
| 2 | 62 | 138 | 8556 |
| 3 | 101 | 146 | 14746 |
| 4 | 67 | 146 | 9782 |
| 5 | 63 | 139 | 8757 |
| 6 | 130 | 150 | 19500 |
| 7 | 72 | 125 | 9000 |
| 8 | 73 | 149 | 10877 |
| 9 | 72 | 138 | 9936 |
| 10 | 54 | 103 | 5562 |
| 11 | 98 | 160 | 15680 |
| 12 | 56 | 106 | 5936 |
| 13 | 118 | 153 | 18054 |
| 14 | 135 | 100 | 13500 |
| 15 | 74 | 144 | 10656 |
| 16 | 65 | 134 | 8710 |

According to the analyzed results in Tables 8,9 and 10, it can be learned that the static task allocation based on the improved genetic hybrid algorithm presents a significant advantage for the same number of tasks. Compared with the traditional genetic algorithm, the method reduced the fleet path cost by about 31.22%,31.0%, and 21.4%, respectively. In addition, the time cost decreased by 18.68%,19.40% and 13.74%. Notably, there was no significant increase in fuel consumption as the number of farm machines increased. The experimental results clearly show that the static task allocation method based on the improved genetic hybrid algorithm performs better when the number of tasks remains constant.

To verify the reasonableness of the algorithm's cost reduction, a series of experiments were conducted on the traditional genetic algorithm and the improved genetic hybrid algorithm in the case of task allocation using 4,5 and 6 farm machines. Each set of experiments was repeated 10 times and averaged to obtain data on average distance cost, average fuel consumption and average time spent; the results are shown in Fig.14, Fig.15 and Fig.16.With the same number of iterations and tasks, the experimental results show that the static task allocation results based on the improved genetic hybrid algorithm can be seen to be about 32.34%,28.10% and 21.54% lower on average in terms of swarm path cost compared to those based on the traditional genetic algorithm;

the average fuel consumption is reduced by 0.195%,0.156% and 0.074%, respectively; and the average time spent is by 9.83%,10.82% and 13.02 %. These data indicate that the static task allocation method using the improved genetic hybrid algorithm is more effective when the number of tasks remains constant. The simulation results further show that using the improved genetic hybrid algorithm can obtain better task allocation results while reducing the distance and time travelled by the agricultural machine. Using the enhanced genetic hybrid algorithm for task allocation can significantly reduce the farm machine's burden while preventing the agricultural machine from making unnecessary travelling paths. Therefore, the static task allocation method based on the improved genetic hybrid algorithm is advantageous. These results further validate the effectiveness of our algorithm in reducing cost and optimizing resource utilization.

Table 8 Results of tasking of 4 agricultural machines

| Serial No. | Methods | AM 1 | AM 2 | AM 3 | AM 4 | $\sum s_i/m$ | $\sum c_i/L$ | $\sum t_i/h$ |
|---|---|---|---|---|---|---|---|---|
| 1 | GA | 12→13 | 14→6→9→10→7→11→8→4→15→3→16→5 | 1 | 2 | 6461.30 | 181.81 | 35.61 |
| 2 | IGHOA | 16→13→8→4→11→3→15→2→9→6→12 | 5→10→7 | 1 | 14 | 4444.20 | 181.64 | 28.96 |

Table 9 Results of tasking of 5 agricultural machines

| Serial No. | Methods | AM 1 | AM 2 | AM 3 | AM 4 | AM 5 | $\sum s_i/m$ | $\sum c_i/L$ | $\sum t_i/h$ |
|---|---|---|---|---|---|---|---|---|---|
| 1 | GA | 14 | 13 | 1 | 9→4→11→7→6→5→16→10→2 | 12→8→3→15 | 7206.40 | 181.65 | 39.13 |
| 2 | IGHOA | 12 | 1 | 14 | 16→13 | 9→2→6→7→15→3→11→4→10→5 | 4972.70 | 181.69 | 31.54 |

Table 10 Results of tasking of 6 agricultural machines

| Serial No. | Methods | AM 1 | AM 2 | AM 3 | AM 4 | AM 5 | AM 6 | $\sum s_i/m$ | $\sum c_i/L$ | $\sum t_i/h$ |
|---|---|---|---|---|---|---|---|---|---|---|
| 1 | GA | 14 | 1 | 7 | 16→13 | 6 | 9→12→5→10→4→8→15→2→11→3 | 7502.2 | 181.59 | 40.2952 |
| 2 | IGHOA | 14 | 1 | 12 | 6→9→2 | 5→10→7→15→3→11→4→8→13 | 16 | 5897.3 | 181.59 | 34.7590 |

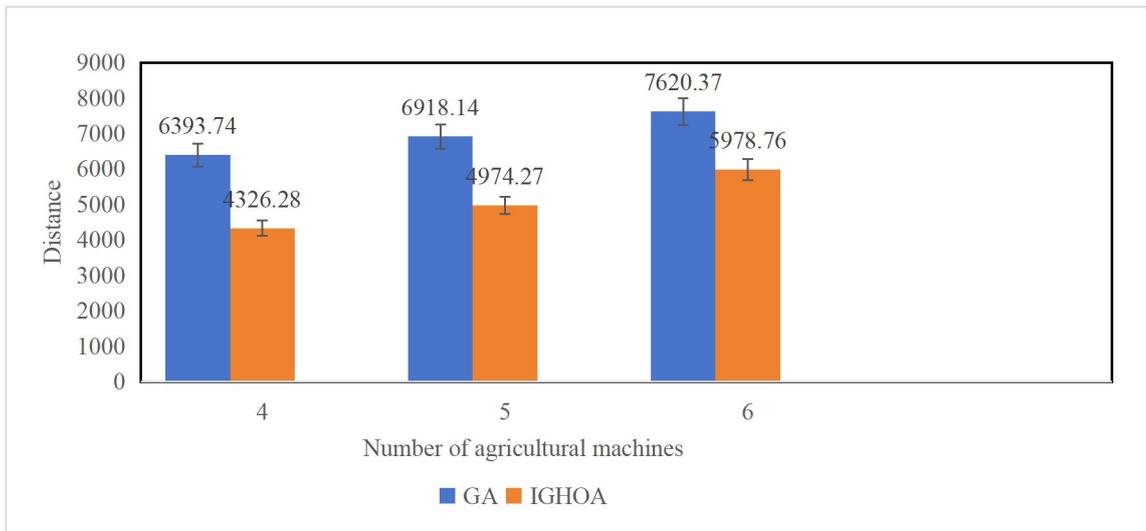

Fig.14 Translation of average distances for different farm machine task assignments

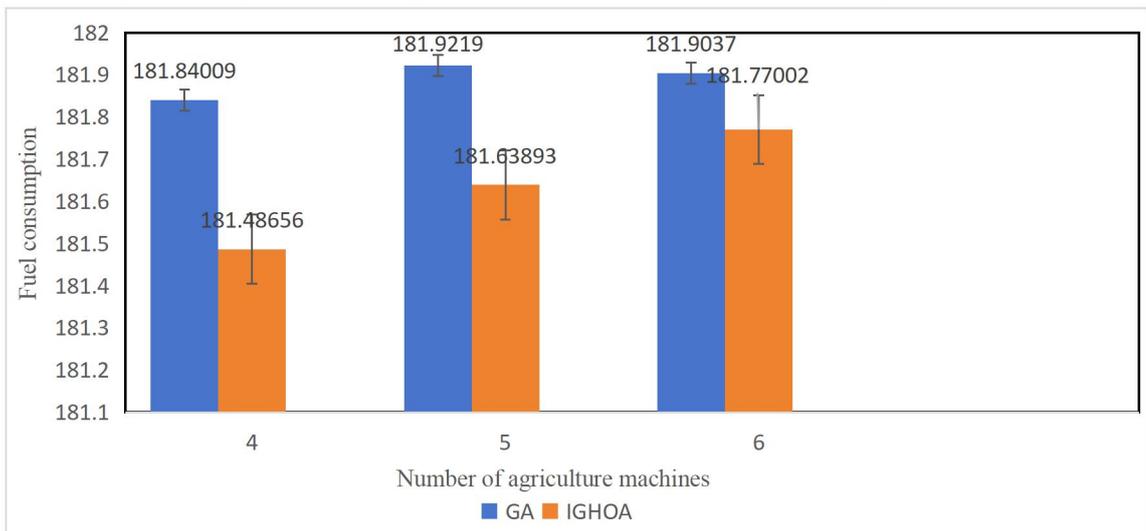

Fig 15 Average fuel consumption for different farm machine task assignments

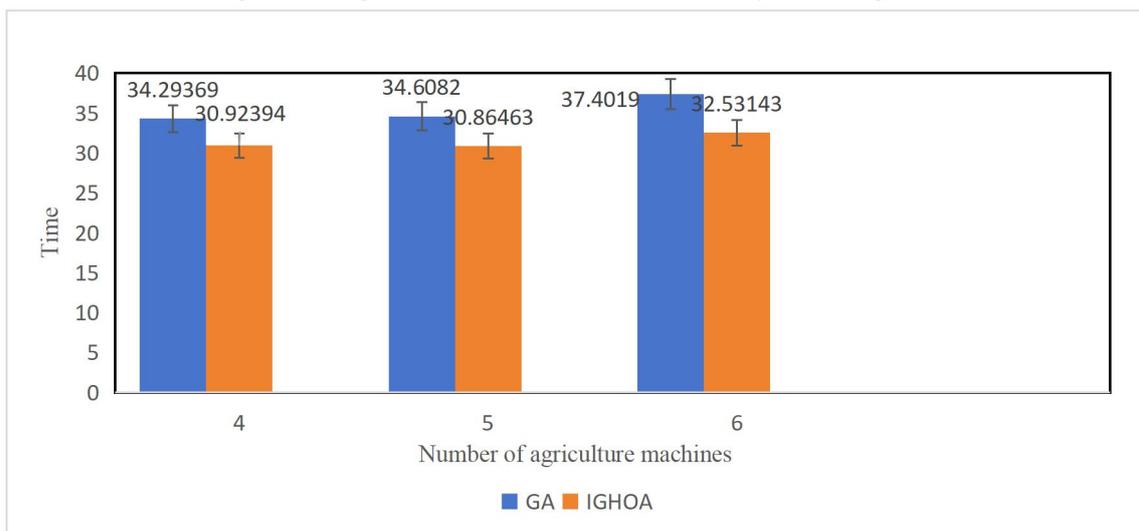

Fig 16 Average time for different farm machinery task assignments

## 5.4 Simulation Experiment on Real-Field Tasks

To validate the effectiveness of the proposed task allocation approach, we conducted algorithmic simulations using real-world farmland coordinates expressed in latitude and longitude. Fig 1 depicts the distribution of operational tasks within Chang ji Huaxing Farm, where 15 duties were assigned using 3 and 6 farm machines. The specific task parameters are detailed in Table 11, sourced from modelling data provided by *Ovitalmap*.Through this simulation experiment, we can more precisely assess the efficacy of the proposed task allocation method in natural farmland settings. This not only enhances the reliability of the investigation but also provides us with a more authentic data-set to validate the algorithm's effectiveness.

Table 11 Parameters of the mission field

| Task number | Widths/m | Lengths/m | Area/m$^2$ |
|---|---|---|---|
| 1 | 241 | 178 | 42898 |
| 2 | 277 | 172 | 47644 |
| 3 | 216 | 166 | 35856 |
| 4 | 214 | 160 | 34240 |
| 5 | 234 | 188 | 43992 |
| 6 | 275 | 183 | 50325 |
| 7 | 212 | 177 | 37524 |
| 8 | 218 | 174 | 37932 |
| 9 | 233 | 153 | 35649 |
| 10 | 273 | 153 | 41769 |
| 11 | 207 | 153 | 31671 |
| 12 | 219 | 153 | 33507 |
| 13 | 307 | 173 | 53111 |
| 14 | 302 | 172 | 51944 |
| 15 | 316 | 157 | 49612 |

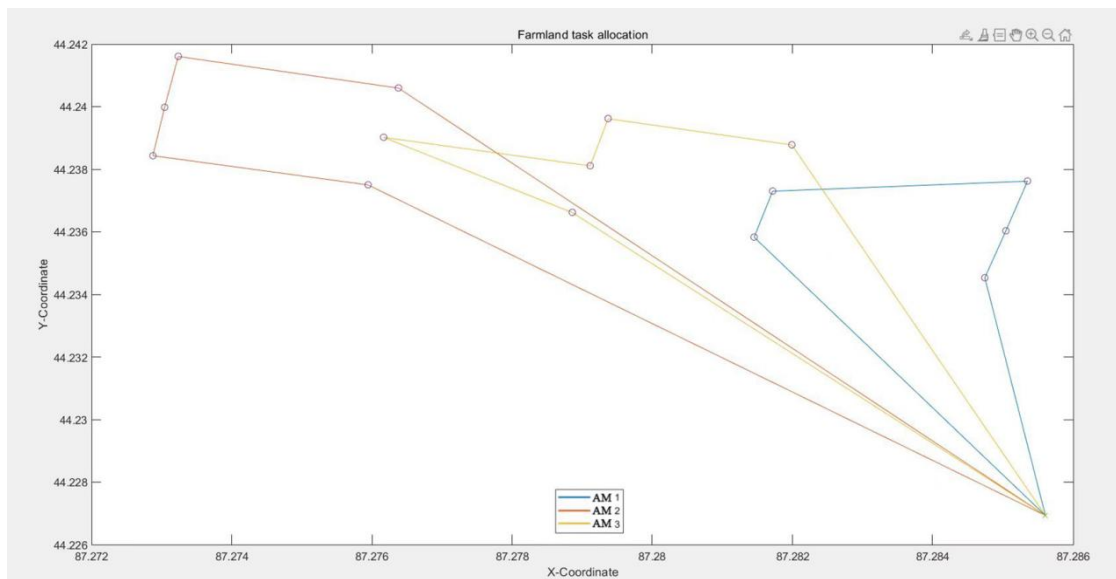

（a）

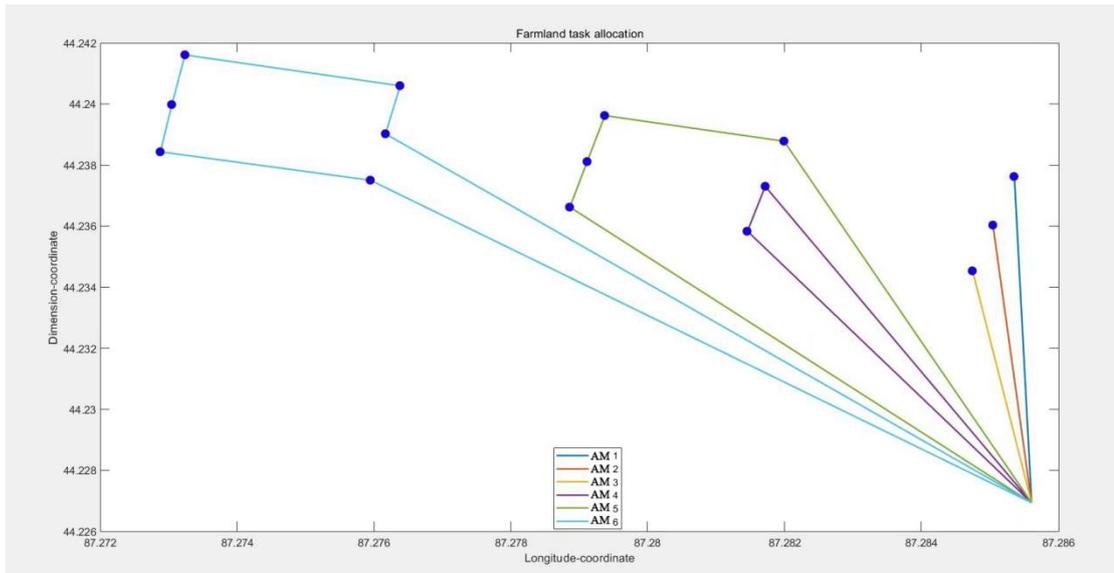

（b）

Fig.17 Simulation results of the proposed method in agricultural fields: (a) 3 Agricultural machines;(b)6 Agricultural machines

The simulation was validated using the farmland's actual latitude and longitude coordinates. We conducted simulation experiments using three and six farm machines, respectively and obtained the results shown in Fig.17.The results show that the method effectively solves the path planning problem used for multiple farm machines.

## 6 Conclusion

This article presents a global path planning and task allocation algorithm. Firstly, we investigated the problem of static task allocation for collaborative multi-agricultural machinery operations. We constructed a model that combines pre-planned paths with static task allocation. Secondly, we introduced logical functions and overshoot and established an adaptive tolerance mechanism that dynamically adjusts crossover operators. In the global search phase, we designed adaptive crossover and adaptive mutation operators to improve the genetic algorithm, enhance population diversity, and better explore the solution space. This approach improves diversity within the population, enabling it to search for new solutions in the search space.

Additionally, we employed the 2-opt local optimization mutation operator and Modified circle algorithm. These methods draw inspiration from local search principles, altering the order of paths or nodes to discover more optimal solutions, playing a crucial role in various optimization algorithms. Finally, we validated the performance of the proposed algorithm using publicly available datasets from TSPLIB. Compared to other methods, we verified the effectiveness of this approach, as it significantly reduces path costs. However, it's important to note that this paper did not consider factors such as obstacles, road conditions, and task allocation fairness in the algorithm. In future work, we will address these issues, including path planning with barriers, road conditions, and balanced task allocation. We will also research how to handle the influence of unknown obstacles on path planning.

**CRediT authorship contribution statement**

Haohao Du: Methodology, Validation, Writing - original draft. Huicheng Lai: Writing - review & editing. Guxue Gao: review & editing. Xiaopeng Wen: review & editing. Yifei Li: review &

editing. Haidong Wang: review & editing. Guo Zhang.

**Declaration of Competing Interest**

The authors declare that they have no known competing financial interests or personal relationships that could have appeared to influence the work reported in this paper.

**Acknowledgments**

This study is supported by the National Key R&D Program of China (Grant Nos. 2022ZD0115803).